# Robust PCA in High-dimension: A Deterministic Approach

**Jiashi Feng** A0066331@NUS.EDU.SG
Department of Electrical and Computer Engineering, National University of Singapore, Singapore

**Huan Xu** MPEXUH@NUS.EDU.SG
Department of Mechanical Engineering, National University of Singapore, Singapore

**Shuicheng Yan** ELEYANS@NUS.EDU.SG
Department of Electrical and Computer Engineering, National University of Singapore, Singapore

## Abstract

We consider principal component analysis for contaminated data-set in the high dimensional regime, where the dimensionality of each observation is comparable or even more than the number of observations. We propose a *deterministic* high-dimensional robust PCA algorithm which inherits all theoretical properties of its *randomized* counterpart, i.e., it is tractable, robust to contaminated points, easily kernelizable, asymptotic consistent and achieves maximal robustness – a breakdown point of 50%. More importantly, the proposed method exhibits significantly better computational efficiency, which makes it suitable for large-scale real applications.

## 1. Introduction

This paper is about robust **p**rincipal **c**omponent **a**nalysis (PCA) for high-dimensional data, a topic that has drawn surging attention in recent years. PCA is one of the most widely used data analysis methods (Pearson, 1901). It constructs a low-dimensional subspace based on a set of principal components (PCs) to approximate the observations in the least-square sense. Standard PCA computes PCs as eigenvectors of the sample covariance matrix. Due to the quadratic error criterion, PCA is notoriously sensitive and fragile, and the quality of its output can suffer severely in the face of even few corrupted samples. Therefore, it is not surprising that many works have been dedicated to robustifying PCA (Hubert et al., 2005; Croux & Ruiz-Gazen, 2005; Candes et al., 2009).

Analyzing high dimensional data – data sets where the dimensionality of each observation is comparable to or even larger than the number of observations – has become a critical task in modern statistics and machine learning (Donoho, 2000). Practical high dimensional data, such as DNA microarray data, financial data, consumer data, and climate data, easily have dimensionality ranging from thousand to billions. Partly due to the fact that extending traditional statistical tools (designed for the low dimensional case) into this high-dimensional regime are often unsuccessful, tremendous research efforts have been made to design fresh statistical tools to cope with such "dimensionality explosion".

The work of Xu et al. (2010a) is among the first to analyze robust PCA algorithms in the high-dimensional setup. They identified three pitfalls, namely diminishing breakdown point, noise explosion and algorithmic intractability, where previous robust PCA algorithms stumble. They then proposed the high-dimensional robust PCA (HR-PCA) algorithm that can effectively overcome these problems, and showed that HR-PCA is tractable, *provably* robust and easily kernelizable. In particular, in contrast to standard PCA and existing robust PCA algorithms, HR-PCA is able to robustly estimate the PCs in the *high-dimensional regime* even in the face of a constant fraction of outliers and extremely low Signal Noise Ratio (SNR) – the breakdown point of HR-PCA is 50%, [1] which is the highest breakdown point can ever be achieved, whereas other existing methods all have breakdown points diminishing to zero. Indeed, to the best of our knowledge, HR-PCA appears to be the only algorithm having these properties in the high-dimensional regime.

[1] Breakdown point is a robustness measure defined as the percentage of corrupted points that can make the output of the algorithm arbitrarily bad.

Briefly speaking, HR-PCA is an iterative method which in each iteration performs standard PCA, and then *randomly* remove *one* point in a way that outliers are more likely to be removed, so that the algorithm converges to a good output. Because in each iteration, only one point is removed, the number of iterations required to find a good solution is at least as much as the number of outliers. This, combined with the fact that PCA is computationally expensive itself, prevents HR-PCA from effectively handling large-scale data-sets with many outliers. In addition, the performance of HR-PCA depends on the ability of the built-in random removal to eliminate outliers correctly, which is only guaranteed in a *probabilistic* manner.

To address these two issues, we propose a *deterministic* high dimensional robust PCA algorithm (DHR-PCA). Specifically, instead of removing *one* point, the proposed algorithm decreases the weights of *all* observations in each iteration, in a way that the total weight of the outliers will decrease faster than that of the true samples. We show that DHR-PCA inherits all desirable theoretical properties of HR-PCA, including tractability, kernelizability, the maximal breakdown point, provable performance guarantee and asymptotical optimality. Moreover, DHR-PCA can be much more computationally efficient than (randomized) HR-PCA. As we show below, the number of iterations for DHR-PCA to converge is nearly constant, in sharp contrast to HR-PCA whose number of iterations required increases linearly with the number of outliers. Simulations in Section 4 show that for any fixed number of iterations, the solution to DHR-PCA is at least as good as HR-PCA, and is significantly better when the number of iterations is small. This is very appealing in practice, as both algorithms are "any-time" algorithms, i.e., one can terminate the algorithms at any time and obtain the best solution so-far.

## 2. Related Work

Besides HR-PCA, there have been abundant works on robust PCA, which we briefly discuss in this section. Robust PCA algorithms focusing on the low-dimensional setup (e.g., Rousseeuw, 1984; Croux & Ruiz-Gazen, 2005; Hubert et al., 2005) can be roughly categorized into two groups. The first group of algorithms pursue robust estimation of the covariance matrix, *e.g.*, *M*-estimator (Maronna, 1976), *S*-estimator (Rousseeuw & Leroy, 1987), and Minimum Covariance Determinant (MCD) estimator (Rousseeuw, 1984). These algorithms generally provide more robust results, but their applicability is severely limited to small or moderate dimensions, as there are not enough observations to robustly estimate a high-dimensional covariance matrix. The second group of algorithms directly maximize certain robust estimation of univariate variance for the projected observations and then obtain maximizers as the candidate principal components (Li & Chen, 1985; Croux & Ruiz-Gazen, 1996; 2005; Hubert et al., 2002). These algorithms inherit the robustness characteristics of the adopted estimators and are qualitatively robust. However, all of these algorithms run into unsolvable issues in the high dimensional regime incurred by the curse of dimensionality as stated in the followings.

The targeted high-dimensional regime poses three main challenges to existing robust PCA methods. First, some robust PCA algorithms have breakdown point inversely proportional to the dimensionality, *e.g.*, *M*-estimator (Maronna, 1976), in the high-dimensional regime their breakdown points will diminish and the results will be arbitrarily bad in presence of even few outliers. Second, widely used outlyingness indicators, including Mahalanobis distance and Stahel-Donoho outlyingness (Donoho, 1982) are no longer valid, due to a phenomenon termed – "noise explosion" (Xu et al., 2010a). This causes the algorithms relying on such outlyingness measures (Hubert et al., 2005) to collapse. The third problem is that the dimensionality may be larger than the number of data points and thus some robust estimators including Minimum Volume Ellipsoid (MVE) and Minimum Covariance Determinant (MCD) (Rousseeuw, 1984) become degenerated. Furthermore, the extremely high computational complexity of these estimators and projection pursuit methods for high dimensional data prevents them from being tractable.

Finally, we discuss recent works addressing robust PCA using low-rank technique. (Candes et al., 2009) developed a framework to perform robust PCA using low-rank matrix decomposition. Yet, their method focuses on the scenario that *random* entries of the observation matrix are arbitrarily corrupted, which differs from our setup where one corrupted data point may change the whole column of the observation matrix. The later setup is then investigated in Xu et al. (2010b). While their proposed method performs well under a small fraction of outliers, it breaks down for larger fraction of outliers – in particular, the breakdown point is far from 50%. Moreover, the performance scales unfavorably with the magnitude of noise, which makes it not suitable for the high-dimensional setup, due to "noise-explosion".

## 3. The Algorithm

In this section, we first formally state the problem setup of the high dimensional robust PCA. Then we provide the details of the proposed DHR-PCA algorithm and finally present the main theoretic results on the performance guarantees of the algorithm.

### 3.1. Problem Setup

In this subsection, we present the formal problem description of PCA for the high dimensional data with contamination. Our setup, detailed below for completeness, largely follows that of Xu et al. (2010a).

Given $n$ observations, there are $t$ observations not corrupted, called authentic samples. The authentic samples $\mathbf{z}_i \in \mathbb{R}^m$ are generated through a linear mapping: $\mathbf{z}_i = A\mathbf{x}_i + \mathbf{n}_i$. Here, noise $\mathbf{n}_i$ is sampled from normal distribution $\mathcal{N}(\mathbf{0}, I_m)$; and the signal $\mathbf{x}_i \in \mathbb{R}^d$ are i.i.d. samples of a random variable $\mathbf{x}$ with mean zero and variance $I_d$. The matrix $A \in \mathbb{R}^{m\times d}$ and the distribution $\mu$ of $\mathbf{x}$ are unknown. We assume $\mu$ is absolutely continuous w.r.t. the Borel measure and spherically symmetric. And $\mu$ has light tails, i.e., there exist constants $K, C > 0$ such that $\Pr(\|\mathbf{x}\| \geq x) \leq K\exp(-Cx)$ for all $x \geq 0$. We are interested in the case where $n \approx m \gg d$, i.e., the dimensionality of observations is much larger than that of signals and of the same order as the number of observations.

The outliers (the corrupted data) are denoted as $\mathbf{o}_1, \ldots, \mathbf{o}_{n-t} \in \mathbb{R}^m$ and they are with arbitrary values. We only require that $n-t \leq t$, i.e., the number of outliers are not more than that of authentic samples. Let $\lambda \triangleq (n-t)/n$ be the fraction of corrupted points. We observe the contaminated dataset

$$\mathcal{Y} \triangleq \{\mathbf{y}_1, \ldots, \mathbf{y}_n\} = \{\mathbf{z}_1, \ldots, \mathbf{z}_t\} \bigcup \{\mathbf{o}_1, \ldots, \mathbf{o}_{n-t}\},$$

and aim to recover the principal components of $A$, *i.e.*, the top eigenvectors $\bar{\mathbf{w}}_1, \ldots, \bar{\mathbf{w}}_d$ of $AA^T$. That is, we seek a collection of orthogonal vectors $\mathbf{w}_1, \ldots, \mathbf{w}_d$, that maximize the following performance metric called the *Expressed Variance* (E.V.):

$$\text{E.V.}(\mathbf{w}_1, \ldots, \mathbf{w}_d) \triangleq \frac{\sum_{j=1}^d \mathbf{w}_j^T AA^T \mathbf{w}_j}{\sum_{j=1}^d \bar{\mathbf{w}}_j^T AA^T \bar{\mathbf{w}}_j}.$$

The E.V. represents the portion of signal $A\mathbf{x}$ being expressed by $\mathbf{w}_1, \ldots, \mathbf{w}_d$. Thus, $1 - \text{E.V.}$ is the reconstruction error of the signal. The E.V. is a commonly used evaluation metric for the PCA algorithms (Xu et al., 2010a; d'Aspremont et al., 2008). It is always less than one, with equality achieved by a perfect recovery, i.e., the vectors $\mathbf{w}_1, \ldots, \mathbf{w}_d$ have the same span as the true principal components $\{\bar{\mathbf{w}}_1, \ldots, \bar{\mathbf{w}}_d\}$.

The distribution $\mu$ affects the performance of the algorithms through its tail. We hence adapt the following tail weight function $\mathcal{V} : [0,1] \to [0,1]$ from Xu et al. (2010a), which essentially represents how the tail of $\bar{\mu}$ contributes to its variance,

$$\mathcal{V}(\alpha) \triangleq \int_{-c_\alpha}^{c_\alpha} x^2 \bar{\mu}(dx),$$

where $\bar{\mu}$ is the one-dimensional margin of $\mu$ and $c_\alpha$ is such that $\bar{\mu}([-c_\alpha, c_\alpha]) = \alpha$. Notice that $\mathcal{V}(0) = 0, \mathcal{V}(1) = 1$, and $\mathcal{V}(\cdot)$ is continuous.

### 3.2. Deterministic HR-PCA Algorithm

Our main algorithm is given in Algorithm 1. Here, a Robust Variance Estimator (RVE) $\bar{V}_{\hat{t}}(\cdot)$ is adopted to identify the candidate principal components. For $\mathbf{w} \in \mathcal{S}_m$, the RVE is defined as $\bar{V}_{\hat{t}}(\mathbf{w}) \triangleq \frac{1}{n}\sum_{i=1}^{\hat{t}} |\mathbf{w}^T\mathbf{y}|^2_{(i)}$, where the subscript $(\cdot)$ denotes a non-decreasing order of the variables. And it can be seen that the RVE stands for the following statistics: project $\mathbf{y}_i$ onto the direction $\mathbf{w}$, replace the furthest $n - \hat{t}$ samples by 0, and then compute the variance. If the variance is large, it is likely that a correct principal component direction is found. Otherwise, a number of points with largest variance may be corrupted. Notice that the RVE is always performed on the original observed set $\mathcal{Y}$. We find that RVE coincides with the robust L-estimator, which is defined as a linear combination of order statistics: $T_n = \sum_{i=1}^n a_{ni} h(x_{(i)})$ for some function $h$.

We now explain our innovation compared to HR-PCA, and its intuition. In HR-PCA, steps 4 and 5 are replaced by a random removal – the probability $\hat{\mathbf{y}}_i$ being removed is proportional to $\sum_{j=1}^d \left(\mathbf{w}_j^T \hat{\mathbf{y}}_i\right)^2$. It has been shown in Xu et al. (2010a) that *in expectation (and in probability)*, either the number of outliers will decrease faster, or the algorithm will find a good solution. Since in each iteration, only one point is removed, the number of iterations required to find a satisfactory output depends linearly on the number of outliers.

Instead of resorting to a random mechanism, DHR-PCA deterministically reduce the effect of corrupted data points. In particular, Moreover, DHR-PCA operates on all the data points in each iteration, which decouples the dependence of the computational cost on the number of outliers and enhances the efficiency significantly compared with HR-PCA. We consider an artificial example to illustrate this: assume both HR-PCA and DHR-PCA requires $M$ iterations for a data-set $\mathcal{Y}_0$. Now suppose a new data-set $\mathcal{Y}$ contains $J$ identical copies of data-set $\mathcal{Y}_0$. Then the number of iterations for DHR-PCA remains unchanged, while HR-PCA requires $JM$ iterations. Simulation results

**Algorithm 1** DHR-PCA.

**Input:** Contaminated sample set $\mathcal{Y} = \{\mathbf{y}_1, \ldots, \mathbf{y}_n\} \subset \mathbb{R}^m$, parameters $d, \hat{t}$.
**Output:** Recovered PCs: $\mathbf{w}_1^*, \ldots, \mathbf{w}_d^*$.
**Initialize** $\hat{\mathbf{y}}_i := \mathbf{y}_i, \alpha_i = 1, \forall i = 1, \ldots, n; \mathrm{Opt} := 0$.
**repeat**

1. Compute the empirical variance matrix
$$\hat{\Sigma} := \frac{1}{n} \sum_{i=1}^{n} \alpha_i \hat{\mathbf{y}}_i \hat{\mathbf{y}}_i^T;$$
2. Perform PCA on $\hat{\Sigma}$. Let $\mathbf{w}_1, \ldots, \mathbf{w}_d$ be the $d$ principle components of $\hat{\Sigma}$;
3. If $\sum_{j=1}^{d} \bar{V}_{\hat{t}}(\mathbf{w}_j) > \mathrm{Opt}$, then let $\mathrm{Opt} := \sum_{j=1}^{d} \bar{V}_{\hat{t}}(\mathbf{w}_j)$ and let $\mathbf{w}_j^* := \mathbf{w}_j$ for $j = 1, \ldots, d$;
4. Calculate
$$\eta = \min_i \frac{1}{\sum_{j=1}^{d} \left(\mathbf{w}_j^T \hat{\mathbf{y}}_i\right)^2}, \forall i : \alpha_i \neq 0.$$
5. Update the sample weight $\alpha_i := \alpha_i - \Delta\alpha_i$, $\forall i : \alpha_i \neq 0$, where $\Delta\alpha_i = \eta \alpha_i \sum_{j=1}^{d} \left(\mathbf{w}_j^T \hat{\mathbf{y}}_i\right)^2$;

**until** Convergence

for more realistic setups, reported in Section 4, also demonstrate that the deterministic algorithm provides higher efficiency than HR-PCA.

Theorem 1 and Theorem 2 below show that the proposed algorithm achieves the same performance guarantees as HR-PCA. The proofs are shown in Section 5.

**Theorem 1.** (Finite Sample Performance) *Let the Algorithm 1 output $\{\mathbf{w}_1, \ldots, \mathbf{w}_d\}$. Fix a $\kappa > 0$, and let $\tau = \max(m/n, 1)$. There exists a universal constant $c_0$ and a constant $C$ which can possibly depend on $\hat{t}/t, \lambda, d, \mu$ and $\kappa$, such that for any $\gamma < 1$, if $n/\log^4 n \geq \log^6(1/\gamma)$, then with probability $1-\gamma$ the following holds*

$$\begin{aligned} &\mathrm{E.V.}\{\mathbf{w}_1, \ldots, \mathbf{w}_d\} \\ \geq\ & \left[\frac{\mathcal{V}\left(1 - \frac{\lambda(1+\kappa)}{(1-\lambda)\kappa}\right)}{(1+\kappa)}\right] \times \left[\frac{\mathcal{V}\left(\frac{\hat{t}}{t} - \frac{\lambda}{1-\lambda}\right)}{\mathcal{V}\left(\frac{\hat{t}}{t}\right)}\right] \\ -\ & \left[\frac{8\sqrt{c_0 \tau d}}{\mathcal{V}\left(\frac{\hat{t}}{t}\right)}\right] \left(\mathrm{trace}(AA^T)\right)^{-1/2} \\ -\ & \left[\frac{2c_0\tau}{\mathcal{V}\left(\frac{\hat{t}}{t}\right)}\right] \left(\mathrm{trace}(AA^T)\right)^{-1} - C\frac{\log^2 n \log^3(1/\gamma)}{\sqrt{n}}. \end{aligned}$$

We also consider the asymptotic performance of the proposed algorithm when the dimension and the number of data points grow together to infinity. Our asymptotic setting is similar to (Xu et al., 2010a). Suppose there exists a sequence of sample sets $\{\mathcal{Y}(j)\} = \{\mathcal{Y}(1), \mathcal{Y}(2), \ldots\}$, where $\mathcal{Y}(j), n(j), m(j), A(j), d(j)$, etc., denote the corresponding values of the quantities defined above, the following must hold for some positive constants $c_1, c_2$:

$$\begin{aligned} &\lim_{j\to\infty} \frac{n(j)}{m(j)} = c_1; d(j) \leq c_2; m(j) \uparrow +\infty; \\ &\mathrm{trace}\left(A(j)^T A(j)\right) \uparrow \infty. \end{aligned} \tag{1}$$

While $\mathrm{trace}\left(A(j)^T A(j)\right) \uparrow \infty$, if it scales slowly than $\sqrt{m(j)}$, the SNR will asymptotically decrease to zero.

The last three terms in Theorem 1 go to zero as the dimension and number of points scale to infinity, i.e., as $n$ and $m \to \infty$. Therefore, we immediately obtain:

**Theorem 2.** (Asymptotic Performance) *Given a sequence of $\{\mathcal{Y}(j)\}$, if the asymptotic scaling in Expression (1) holds, and $\limsup \lambda(j) \leq \lambda^*$, then the following holds in probability when $j \uparrow \infty$ (i.e., when $n$ and $m \uparrow \infty$),*

$$\begin{aligned} &\liminf_j \mathrm{E.V.}\{\mathbf{w}_1(j), \ldots, \mathbf{w}_d(j)\} \\ \geq\ & \max_\kappa \left[\frac{\mathcal{V}\left(1 - \frac{\lambda^*(1+\kappa)}{(1-\lambda^*)\kappa}\right)}{(1+\kappa)}\right] \times \left[\frac{\mathcal{V}\left(\frac{\hat{t}}{t} - \frac{\lambda^*}{1-\lambda^*}\right)}{\mathcal{V}\left(\frac{\hat{t}}{t}\right)}\right] \end{aligned} \tag{2}$$

Observe that when $\lambda^* = 0$, i.e., the number of outliers scales sublinearly, the right-hand-side converges to 1 by taking $\kappa(j) = \sqrt{\lambda(j)}$, implying that the algorithm is asymptotically optimal. On the other hand, for any $\lambda < 0.5$, the right hand side is strictly positive (picking $\kappa$ large enough), implying that the breakdown point converges to 50%.

For small $\lambda$, we can make use of the light tail condition on $\bar{\mu}$, to establish the following bound that simplifies (2). The proof is deferred to the supplementary material.

**Corollary 1.** *Under the settings of the above theorem, the following holds in probability when $j \uparrow \infty$ (i.e., when $n, p \uparrow \infty$),*

$$\liminf_j \mathrm{E.V.}\{\mathbf{w}_1(j), \ldots, \mathbf{w}_d(j)\} \geq 1 - \frac{C'\sqrt{\alpha\lambda^* \log(1/\lambda^*)}}{\mathcal{V}(0.5)}.$$

Before concluding this section, we remark that DHR-PCA is easily kernelizable. Specifically, given a mapping function $\phi(\cdot) : \mathbb{R}^m \to \mathcal{H}$ and kernel function $k(\cdot, \cdot)$ satisfying $k(\mathbf{x}, \mathbf{y}) = \langle \phi(\mathbf{x}), \phi(\mathbf{y}) \rangle$ for all $\mathbf{x}, \mathbf{y} \in \mathbb{R}^m$, we can perform dimension reduction without requiring the explicit form of $\phi(\cdot)$ in the kernel PCA (Schölkopf et al., 1997). In particular, for the centered mapped features $\{\phi(\mathbf{y}_1), \cdots, \phi(\mathbf{y}_n)\}$, the output PCs can be represented as

$$\mathbf{w}_q = \sum_{j=1}^{n} a_j(k)\phi(\hat{\mathbf{y}}_j).$$

And the feature projection can be calculated by

$$\langle \mathbf{w}_q, \phi(\mathbf{v}) \rangle = \sum_{j=1}^{n} a_j(q) k(\hat{\mathbf{y}}_j, \mathbf{v}),$$

where $a(q)$ is the $q^{th}$ eigenvector of the kernel matrix. Note that Algorithm 1 only involves calculating $\langle \mathbf{w}_q, \phi(\mathbf{y}_i) \rangle$ (in RVE evaluation) and $\langle \mathbf{w}_q, \phi(\sqrt{\alpha_i}\mathbf{y}_i) \rangle$ (in decreasing values of $\alpha_i$'s). Since the kernelization of both these two steps are obtained, the DHR-PCA algorithm can be kernelized easily.

## 4. Simulations

We devote this section to experimentally comparing the proposed DHR-PCA with HR-PCA. Since HR-PCA has shown superior robustness (against the dimensionality and number of outliers) over several robust PCA algorithms and standard PCA (Xu et al., 2010a), we skip simulations for them here.

The numerical study is aimed to illustrate that DHR-PCA is much more efficient than HR-PCA, and meanwhile it achieves competitive performance. Here, we report the results for $d = 1$. We follow the data generation method in (Xu et al., 2010a) to randomly generate an $m \times 1$ matrix and then scale its leading singular value to $\sigma$. A $\lambda$ fraction of outliers are generated on a line with a uniform distribution over $[-\sigma \cdot \text{mag}, \sigma \cdot \text{mag}]$. Thus, "mag" represents the ratio between the magnitude of the outliers and that of the signal $A\mathbf{x}_i$ and is fixed as 10. The value of $\hat{t}$ is set as $(1 - \lambda)n$, if $\lambda$ is known exactly. Otherwise, $\hat{t}$ can be simply set as $0.5n$. For each parameter setup, we report the average result of 20 tests and standard deviation.

Figure 4 shows the results for $m = 100$, 1000 and 10000 cases respectively with $\sigma = 5$. From the figure, we can make following observations. Firstly, DHR-PCA converges much faster than HR-PCA, especially for a large number of outliers. For example, when $m = 10000$ and $\lambda = 0.4$, the proposed algorithm converges using less than 2 iterations in average while HR-PCA needs more than 4000 iterations to converge. Secondly, the computational time for DHR-PCA in *each* iteration is always in the same order as HR-PCA. These results well demonstrate that DHR-PCA is much more efficient than HR-PCA.

As for the performance, i.e., the E.V. of the recovered PCs, Figure 4 shows that DHR-PCA performs competitively to HR-PCA. For all the cases, the E.V. of final solution of DHR-PCA is always larger than that of HR-PCA. Moreover, if we terminate both algorithms at any early iteration, DHR-PCA always perform better than HR-PCA. This is appealing in practice, as we can terminate DHR-PCA at any time and obtain a satisfactory result in practical implementation. In addition, both DHR-PCA and HR-PCA perform quite well even in presence of varying number of outliers ($\lambda = 0.05$ to 0.4) and small signal magnitude ($\sigma = 5$), which coincides with the results in (Xu et al., 2010a).

We then investigate the relationship between the number of iterations before convergence and the number of outliers for the two methods. As shown in Figure 2, the number of iterations taken by HR-PCA is approximately proportional to the number of corrupted points. This is not surprising, since in each iteration HR-PCA removes at most one outlier. In a stark contrast, the number of required iterations of DHR-PCA remains nearly constant, shown by the flat curve in the figures. This demonstrates that DHR-PCA has good scalability and can potentially be applied to large real applications. We provide more simulations under numerous settings in the appendix.

## 5. Proof of Theorem 1

In this section, we sketch the proof of Theorem 1. In what follows, we let $d, m/n, \lambda, \hat{t}/t$, and $\mu$ be fixed. We can fix a $\lambda \in (0, 0.5)$ w.l.o.g. due to the fact that if a result is shown to hold for $\lambda$, then it holds for $\lambda' < \lambda$. The letter $c$ is used to represent a constant, and $\epsilon$ is a constant that decreases to zero as $n$ and $m$ increase to infinity. Let $\mathbf{w}_1(s), \ldots, \mathbf{w}_d(s)$ be the candidate solution at stage $s$. Let $\mathcal{Z}$ and $\mathcal{O}$ be the sets of indices of authentic samples and corrupted samples respectively. We let $\mathcal{B}_d \triangleq \{\mathbf{w} \in \mathbb{R}^d | \|\mathbf{w}\| \leq 1\}$, and $\mathcal{S}_d$ be its boundary. Here Theorems 3 and 4 are directly adapted from (Xu et al., 2010a).

### 5.1. Validity of the Robust Variance Estimator

We first show that the following condition holds with high probability. The detailed proof can be found in (Xu et al., 2010a).

**Condition 1.** *There exists* $\epsilon_1, \epsilon_2, \bar{c}$ *such that* *(I)*$\sup_{\mathbf{w} \in \mathcal{S}_d} \left| \frac{1}{t} \sum_{i=1}^{t'} \left| \mathbf{w}^T \mathbf{x} \right|^2_{(i)} - \mathcal{V}\left(\frac{t'}{t}\right) \right| \leq \epsilon_1;$

*(II)* $\sup_{\mathbf{w}\in\mathcal{S}_d} \left|\frac{1}{t}\sum_{i=1}^{t} \left|\mathbf{w}^T\mathbf{x}_i\right|^2 - 1\right| \leq \epsilon_2$; *(III)* $\sup_{\mathbf{w}\in\mathcal{S}_m} \frac{1}{t}\sum_{i=1}^{t} \left|\mathbf{w}^T\mathbf{n}_i\right|^2 \leq \bar{c}$.

**Theorem 3.** *Fix any $\eta < 1$. With probability at least $1 - 3\gamma$, Condition 1 holds uniformly for all $t' \leq \eta t$, with $\bar{c} = c\tau(1 + \frac{\log(1/\gamma)}{n})$, $\epsilon_2 = c\log^2 n \log^3(1/\gamma)/\sqrt{n}$, and $\epsilon_1 = c\sqrt{\frac{\log n + \log(1/\gamma)}{n}} + \frac{c\log^{2.5} n \log^{3.5}(1/\gamma)}{n}$, for a constant $c$ possibly depends on $d$, $\mu$ and $\eta$.*

Under Condition 1, RVE is a good estimator.

**Theorem 4.** *Let $t' \leq t$. Suppose Condition 1 holds. Then for all $\mathbf{w} \in \mathcal{S}_m$ the following holds:*

$$\begin{aligned} &(1-\epsilon_1)\|\mathbf{w}^T A\|^2 \mathcal{V}\left(\frac{t'}{t}\right) - 2\|\mathbf{w}^T A\|\sqrt{(1+\epsilon_2)\bar{c}} \\ \leq\ & \frac{1}{t}\sum_{i=1}^{t'} |\mathbf{w}^T\mathbf{z}|^2_{(i)} \\ \leq\ & (1+\epsilon_1)\|\mathbf{w}^T A\|^2 \mathcal{V}\left(\frac{t'}{t}\right) + 2\|\mathbf{w}^T A\|\sqrt{(1+\epsilon_2)\bar{c}} + \bar{c}. \end{aligned}$$

From the above theorem, we can immediately obtain the following corollary.

**Corollary 2.** *Let $t' \leq t$. Suppose Condition 1 holds. Then for all any $\mathbf{w}_1, \cdots, \mathbf{w}_d \in \mathcal{S}_m$ the following holds:*

$$\begin{aligned} &(1-\epsilon_1)\mathcal{V}\left(\frac{t'}{t}\right) H(\mathbf{w}) - 2\sqrt{(1+\epsilon_2)\bar{c} d H(\mathbf{w})} \\ \leq\ & \sum_{j=1}^{d}\frac{1}{t}\sum_{i=1}^{t'} |\mathbf{w}_j^T\mathbf{z}|^2_{(i)} \\ \leq\ & (1+\epsilon_1)\mathcal{V}\left(\frac{t'}{t}\right) H(\mathbf{w}) + 2\sqrt{(1+\epsilon_2)\bar{c} d H(\mathbf{w})} + \bar{c}, \end{aligned}$$

*and*

$$\begin{aligned} &(1-\epsilon)H(\mathbf{w}) - 2\sqrt{(1+\epsilon)\bar{c} d H(\mathbf{w})} \\ \leq\ & \sum_{j=1}^{d}\frac{1}{t}\sum_{i=1}^{t} |\mathbf{w}_j^T\mathbf{z}_i|^2 \\ \leq\ & (1+\epsilon)H(\mathbf{w}) + 2\sqrt{(1+\epsilon)\bar{c} d H(\mathbf{w})} + \bar{c}, \end{aligned}$$

*where $H(\mathbf{w}) \triangleq \sum_{j=1}^{d} \|\mathbf{w}_j^T A\|^2$.*

### 5.2. Finite Steps for a Good Solution

In this step, we show that the algorithm finds a good solution in a small number of steps. Proving this involves showing that at any given step, either the algorithm finds a good solution, or the weight adjusting step decreases weights of corrupted points more than the authentic points. Let $\alpha_i^{(s)}$ denote the weight of the $i^{th}$ data point in the $s^{th}$ stage. These points are a good solution if the variance of the points projected onto their span is mainly due to the authentic samples rather than the corrupted points. We denote this "good output event at step $s$" by $\mathcal{E}(s)$, defined as:

$$\mathcal{E}(s) = \left\{\sum_{i\in\mathcal{Z}} \alpha_i^{(s)} v_i(s) \geq \frac{1}{\kappa}\sum_{i\in\mathcal{O}} \alpha_i^{(s)} v_i(s)\right\},$$

where the variance $v_i(s) = \sum_{j=1}^{d}\left(\mathbf{w}_j(s)^T\mathbf{y}_i\right)^2$. The intuition is that there cannot be too many steps without finding a good solution, since too many weights of the corrupted points will have been decreased to zero.

**Theorem 5.** *The event $\mathcal{E}(s)$ is true for some $1 \leq s \leq s_0$, where $s_0 \leq \frac{\lambda n(1+\kappa)}{\kappa}$.*

The proof of the above theorem is provided in the supplementary material. We compare Theorem 5 with its randomized counterpart, Theorem 9 of Xu et al. (2010a). The latter states that for HR-PCA, $\mathcal{E}(s)$ succeeds with high probability for some $s \leq (1+\epsilon)(1+\kappa)\lambda n/\kappa$, where $\epsilon$ depends on $\kappa$ and $\lambda$, and decreases to 0 when $n \uparrow \infty$ (for fixed $\kappa$ and $\lambda$). Thus, the advantage of Theorem 5 is two-fold: it is deterministic as opposed to probabilistic, and it does not require the decreasing $\epsilon$.

### 5.3. Bounds on the Solution Performance

Let $\bar{\mathbf{w}}_1, \ldots, \bar{\mathbf{w}}_d$ be the eigenvectors corresponding to the $d$ largest eigenvalues of $AA^T$, namely the optimal solution, $\mathbf{w}_1^*, \ldots, \mathbf{w}_d^*$ be the output of the Algorithm 1 and $\mathbf{w}_1(s), \ldots, \mathbf{w}_d(s)$ be the candidate solution at stage $s$. We define $H(\mathbf{w}_1, \ldots, \mathbf{w}_d) \triangleq \sum_{j=1}^{d} \|\mathbf{w}_j^T A\|^2$, and for notational simplification, let $\bar{H} \triangleq H(\bar{\mathbf{w}}_1, \ldots, \bar{\mathbf{w}}_d)$, $H_s \triangleq H(\mathbf{w}_1(s), \ldots, \mathbf{w}_d(s))$, and $H^* \triangleq H(\mathbf{w}_1^*, \ldots, \mathbf{w}_d^*)$.

The statement of the finite-sample and asymptotic theorems (Theorem 1 and Theorem 2, respectively) lower bound the expressed variance, E.V., which is the ratio $H^*/\bar{H}$. The final part of the proof accomplishes this in two main steps. First, we lower bound $H_s$ in terms of $\bar{H}$ where $s$ is some step for which $\mathcal{E}(s)$ is true, i.e., the principal components found by the $s^{th}$ step of the algorithm are "good". By Theorem 5, we know that there is a "small" such $s$. Based on the true $\mathcal{E}(s)$ and the algorithm definition, we can conclude the bound via some algebraic manipulations. The final output of the algorithm, however, is only guaranteed to have a high value of the robust variance estimator, $\bar{V}$ - that is, even if there is a "good" solution at some intermediate step $s$, we do not necessarily have a way of identifying it. Thus, the next step lower bounds the value of $H^*$ in terms of the value $H$ of any output $\mathbf{w}_1', \ldots, \mathbf{w}_d'$ that has a smaller value of the robust variance estimator. The details of these two steps are deferred to the supplementary material. Combining the results of above two steps, we can obtain the following theorem providing a lower bound of the ratio $H^*/\bar{H}$, i.e., the expressed variance.

**Theorem 6.** *If $\bigcup_{s=1}^{s_0}\mathcal{E}(s)$ is true, and there exist $\epsilon_1 < 1$, $\epsilon_2$, $\bar{c}$ such that $\sup_{\mathbf{w}\in\mathcal{S}_d}\left|\frac{1}{t}\sum_{i=1}^{t-s_0}|\mathbf{w}^T\mathbf{x}|_{(i)}^2 - \mathcal{V}\left(\frac{t-s_0}{t}\right)\right| \leq \epsilon_1$ and Condition 1 holds, then*

$$\begin{aligned}\frac{H^*}{\bar{H}} \geq\ & \frac{(1-\epsilon_1)^2\mathcal{V}\left(\frac{\hat{t}}{t}-\frac{\lambda}{1-\lambda}\right)\mathcal{V}\left(\frac{t-s_0}{t}\right)}{(1+\epsilon_1)(1+\epsilon_2)(1+\kappa)\mathcal{V}\left(\frac{\hat{t}}{t}\right)} \\ & - \left[\frac{(D_1+D_2)\sqrt{(1+\epsilon_2)\bar{c}d}}{(1+\epsilon_1)(1+\epsilon_2)(1+\kappa)}\right](\bar{H})^{-1/2} \\ & - \left[\frac{(1-\epsilon_1)\mathcal{V}\left(\frac{\hat{t}}{t}-\frac{\lambda}{1-\lambda}\right)\bar{c}+(1+\epsilon_2)\bar{c}}{(1+\epsilon_1)(1+\epsilon_2)\mathcal{V}\left(\frac{\hat{t}}{t}\right)}\right](\bar{H})^{-1},\end{aligned} \tag{3}$$

*where $D_1 = (2\kappa+4)(1-\epsilon_1)\mathcal{V}\left(\frac{\hat{t}}{t}-\frac{\lambda}{1-\lambda}\right)$ and $D_2 = 4(1+\epsilon_2)(1+\kappa)$.*

By bounding all diminishing terms in the right hand side of (3), it reduces to Theorem 1. And Theorem 2 follows immediately. The proofs of Theorem 6 and Theorem 1 are similar to those in (Xu et al., 2010a) and we omit it here.

## 6. Conclusions

In this work, we proposed a deterministic robust PCA algorithm for high-dimensional data corrupted by arbitrary outliers. The algorithm alternates between a classical PCA and decrease of weight coefficients on all the data points. Theoretical analysis showed that the proposed algorithm is tractable, robust to corrupt points, easily kernelizable, asymptotic consistent and achieving maximal breakdown point of 50% – to the best of our knowledge, the first *deterministic* algorithm that achieves these properties in the high-dimensional setup. More importantly, simulation results demonstrated that the proposed algorithm improves computational efficiency over its randomized counterpart HR-PCA – indeed, the number of iterations required to find a satisfactory solution appears to approximate constant, in sharp contrast to HR-PCA whose number of iterations required increases linearly with the number of outliers.

## Acknowledgements

H. Xu is partially supported by National University of Singapore startup grant R-265-000-384-133. S. Yan is partially supported by Singapore Ministry of Education under research Grant MOE2010-T2-1-087.

## References

Candes, E.J., Li, X., Ma, Y., and Wright, J. Robust principal component analysis? *ArXiv:0912.3599*, 2009.

Croux, C. and Ruiz-Gazen, A. A fast algorithm for robust principal components based on projection pursuit. In *COMPSTAT: Proceedings in computational statistics*, 1996.

Croux, C. and Ruiz-Gazen, A. High breakdown estimators for principal components: the projection-pursuit approach revisited. *Journal of Multivariate Analysis*, 2005.

d'Aspremont, A., Bach, F., and Ghaoui, L. Optimal solutions for sparse principal component analysis. *JMLR*, 2008.

Donoho, D.L. Breakdown properties of multivariate location estimators. Technical report, 1982.

Donoho, D.L. High-dimensional data analysis: The curses and blessings of dimensionality. *AMS Math Challenges Lecture*, 2000.

Hubert, M., Rousseeuw, P.J., and Verboven, S. A fast method for robust principal components with applications to chemometrics. *Chemometrics and Intelligent Laboratory Systems*, 2002.

Hubert, M., Rousseeuw, P.J., and Branden, K.V. Robpca: a new approach to robust principal component analysis. *Technometrics*, 2005.

Li, G. and Chen, Z. Projection-pursuit approach to robust dispersion matrices and principal components: primary theory and monte carlo. *Journal of the American Statistical Association*, 1985.

Maronna, R.A. Robust m-estimators of multivariate location and scatter. *The annals of statistics*, 1976.

Pearson, K. On lines and planes of closest fit to systems of points in space. *Philosophical Magazine*, 1901.

Rousseeuw, P.J. Least median of squares regression. *Journal of the American statistical association*, 1984.

Rousseeuw, P.J. and Leroy, A.M. *Robust regression and outlier detection.* John Wiley & Sons Inc, 1987.

Schölkopf, B., Smola, A., and Müller, K.R. Kernel principal component analysis. *Artificial Neural Networks*, 1997.

Xu, H., Caramanis, C., and Mannor, S. Principal component analysis with contaminated data: The high dimensional case. In *COLT*, 2010a.

Xu, H., Caramanis, C., and Sanghavi, S. Robust PCA via outlier pursuit. In *NIPS*, 2010b.

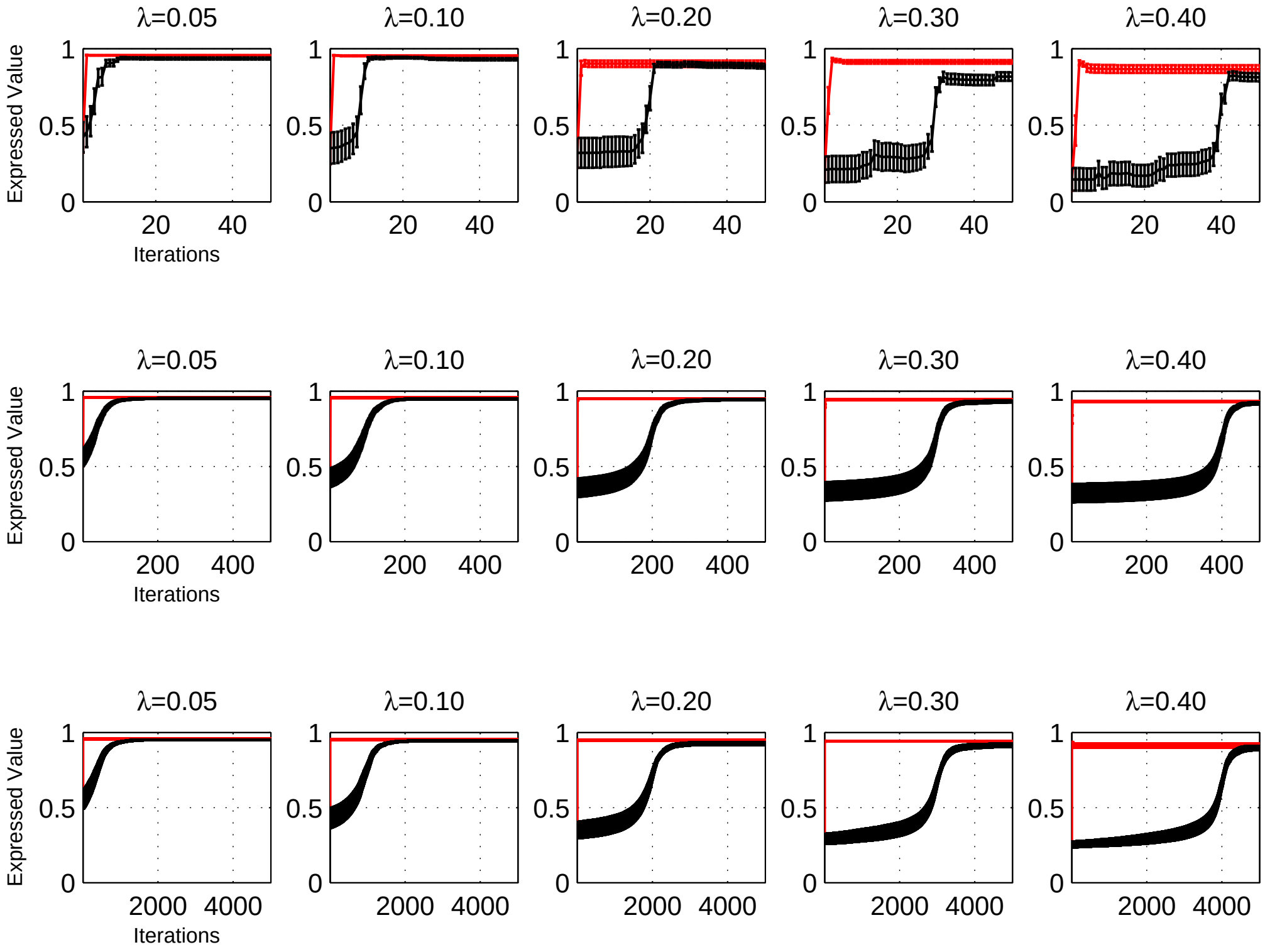

*Figure 1.* DHR-PCA (red line) vs. HR-PCA (black line) with $\sigma = 5$. Upper panel: $m = n = 100$, middle panel: $m = n = 1000$ and bottom panel: $m = n = 10000$. The horizontal axis is the iteration and the vertical axis is the expressive variance value. Please refer to the color version.

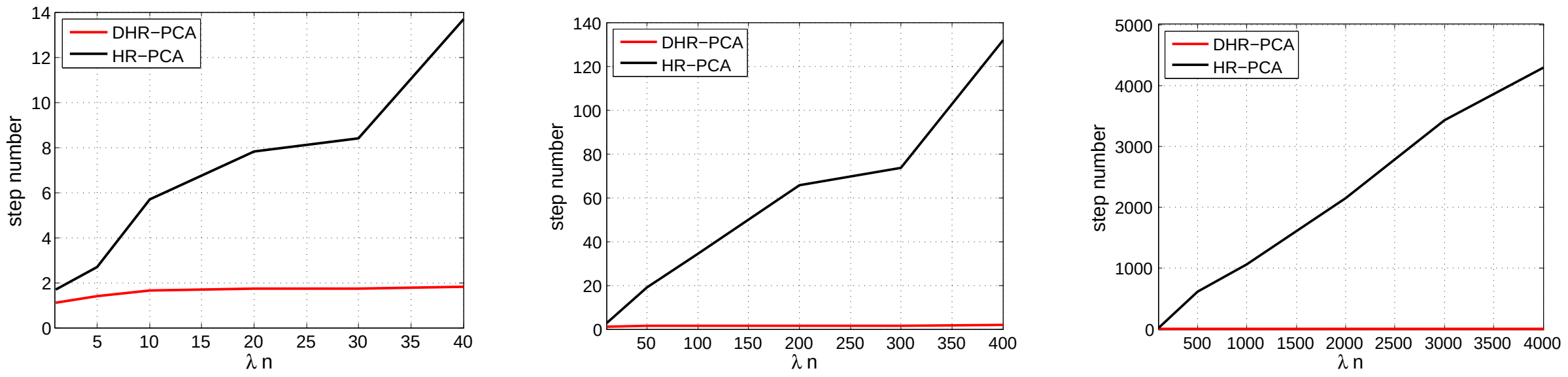

(a) $m = n = 100$ (b) $m = n = 1000$ (c) $m = n = 10000$

*Figure 2.* DHR-PCA (red line) vs. HR-PCA (black line) on the iterative steps taken by them before convergence with $\sigma = 5$ and different dimensionality. The horizontal axis $\lambda n$ is number of corrupted data points and the vertical axis is the number of steps. Please refer to the color version.